\documentclass[sigconf]{acmart}
\AtBeginDocument{%
  }

\setcopyright{acmlicensed}
\copyrightyear{2027}
\acmYear{2027}
\acmDOI{XXXXXXX.XXXXXXX}

\acmConference[WWW '27]{The ACM Web Conference 2027}{May 10--14, 2027}{Dublin, Ireland}
\acmISBN{978-1-4503-XXXX-X/2027/05}

\usepackage{mathrsfs}
\usepackage{dsfont} 
\usepackage[T1]{fontenc}
\usepackage[table,xcdraw]{xcolor}
\usepackage{listings}
\usepackage{colortbl}
\usepackage{xurl}
\usepackage{fancyvrb}
\usepackage{float}

\usepackage{amssymb,amsmath}
\usepackage[normalem]{ulem}
\usepackage{xspace}

\usepackage{makecell}

\newfloat{listing}{htbp}{lop}[section]
\floatname{listing}{Listing}

\usepackage{booktabs}
\usepackage{longtable}
\usepackage{array}
\usepackage{xltabular}
\usepackage{nicefrac}
\usepackage{siunitx}
\usepackage{array,framed}
\usepackage{booktabs}
\usepackage{tabularx}
\usepackage{kotex}
\usepackage{multirow}
\usepackage{multicol}

\usepackage{float, wrapfig, graphicx, subcaption}
\usepackage{tikz}
\usepackage{pifont}

\definecolor {red}{RGB}{210, 80, 80}
\definecolor{green}{RGB}{80, 180, 80}

\usepackage{rotating}
\usepackage{ulem}
\usepackage{multirow,pifont}
\usepackage{fontawesome5}

\newcommand{\cmark}{\ding{51}}
\newcommand{\xmark}{\ding{55}}

\newcommand{\ib}{\textit{SBIA}\xspace}

\newcommand{\pass}{\textcolor{green!60!black}{\faCheck}}
\newcommand{\flag}{\textcolor{red}{\faTimes}}

\begin{document}

\title{MEMORY Wins All: Stealthy Bias Injection Attacks\\via Social Media Content}

\author{Minjae Seo$^{1,4}$, Wonwoo Choi$^2$, Geonwoo Han$^3$, Taekyoung Kwon$^4$, Yongsu Kim$^1$, Sang Seo$^1$,\\ Jaewon Noh$^1$, Hankyul Baek$^1$, Seongyun Seo$^1$, and Myoungsung You$^3$}
\affiliation{%
  \vspace{0.1cm}
  \institution{$^1$ETRI, $^2$ADD, $^3$University of Seoul, $^4$Seoul National University}
  \country{}%
}
\renewcommand{\shortauthors}{Seo et al.}

\begin{abstract}


Personal AI agents consume external content such as web pages and SNS feeds, and retain selected information in persistent memory. As a result, external content has become a new attack vector for memory poisoning, which steers an agent's later actions toward an adversary's bias. However, most existing attacks embed a malicious instruction in external content to make the agent retain the poisoned information, and prompt-injection filters can detect such content. We observe that agents access external content primarily through skills while curating a subset of the content into their memory. Leveraging this curation path, we present \ib{}, a Stealthy Bias Injection Attack that introduces adversarial bias through ordinary external content while relying on the agent itself to perform the memory write. The adversary posts crafted comments under hot SNS feeds and distributes an SNS feed summarization skill whose curation rule admits these comments into memory. Comment cloaking expresses the bias as an ordinary opinion rather than an instruction, comment watermarking allows the skill to identify the crafted comments, and topic anchoring records them so that later related requests recall them. 
On \texttt{BiasBench} with 7,000 crafted SNS comments and seven LLMs on OpenClaw, \ib{} achieves attack success rates of at least 96\% across models and even 91\% on GPT-5.5 in four downstream tasks. Despite this effectiveness, its skill passes all eight evaluated malicious-skill scanners and ClawHub's official skill scanning pipeline. We further propose a memory boundary defense that detects 82.1\% of injected entries at a 5.6\% FPR. Our findings show that external Web content and third-party skill marketplaces form a new attack surface for personal AI agent memory.

\end{abstract}

\maketitle

\section{Introduction}
\label{sec:introduction}

\begin{figure}[t]
    \centering
    \includegraphics[width=1\linewidth]{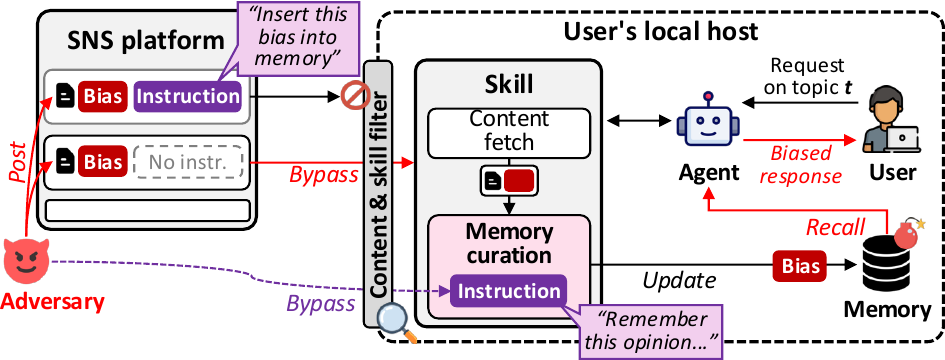}
    \caption{The overview of \ib{} workflow.}
    \label{fig:IBIA_intro}
    \vspace{-2mm}
\end{figure}

Personal AI agents increasingly automate routine tasks beyond simple question answering and one-off queries. Platforms such as OpenClaw~\cite{openclaw}, Hermes Agent~\cite{hermes}, and NemoClaw~\cite{nemoclaw} allow users to delegate email processing, web browsing, document drafting, and SNS feed summarization. Operating on a user's local host with limited human intervention, these agents routinely consume external content from emails, web pages, and social feeds.

To maintain context across sessions, agents retain selected information and task execution results in persistent memory. Recalled memory records are incorporated into subsequent inference contexts as trusted context, allowing them to influence agent behavior across future requests. This also creates a security risk when biased information from external sources enters memory. Specifically, \emph{memory poisoning} occurs when an adversary plants incorrect or biased information on a target topic in the external content (e.g., SNS feeds or emails) that an agent consumes, and causes it to persist in the agent's memory, so that the agent's later actions align with the injected bias~\cite{chen2024agentpoison, zou2025poisonedrag, dong2026memory, srivastava2025memorygraft}. Although prior studies have demonstrated the risk of memory poisoning and proposed various attacks, their attack methods are readily detected or ill-suited to personal AI agents, owing to the following limitations.

\noindent\textbf{Limitations.}
For memory poisoning to succeed, the adversary must induce the agent to retain the injected bias in its memory. Most existing attacks achieve this by embedding an explicit instruction in the external content that the agent reads, prompting it to store the injected bias~\cite{pulipaka2026hidden, zou2026poison, yang2026drunkagent}. Although effective, such an instruction makes the content distinct from benign content, and prompt-injection filters that screen retrieved content readily detect it. For example, an SNS comment that orders the agent to save a given statement in memory is trivially flagged. Attacks that instead deliver the payload through a file carrying executable code~\cite{srivastava2025memorygraft} are likewise caught by the scanners that inspect the skill supply chain. A separate line of work writes the bias into memory directly~\cite{chen2024agentpoison, zou2025poisonedrag, zhang2026memmorph} or submits crafted queries to the agent~\cite{dong2026memory}. However, direct memory writes presume access to the memory store on the user's local host, and crafted queries assume that a single agent serves multiple users. Neither assumption holds for a personalized agent that runs on the local host and serves a single user.

\noindent\textbf{Our approach.} We observe that personal agents access external content primarily through skills, and in this curation path they select a subset of external content and commit it to memory. This provides an opportunity to poison memory stealthily, without embedding any instruction in the external content. Building on this observation, we present \ib{}, a Stealthy Bias Injection Attack that introduces adversarial bias through ordinary external content while relying on the agent itself to perform the memory write. 

\ib{} targets the common scenario in which an agent collects SNS hot feeds and comments, summarizes them, and reports the result to the user. As shown in Figure~\ref{fig:IBIA_intro}, the adversary posts crafted comments expressing a target bias under trending feeds, without embedding any instruction to the agent. When the agent collects and summarizes these feeds, selected comments enter persistent memory through its curation process. When the user later queries the target topic $t$, the agent recalls the retained records and produces a response aligned with the adversary's bias, without the adversary ever accessing the memory or controlling the user's request. To this end, \ib{} applies three techniques to the collected feed. \emph{Comment cloaking} integrates the target bias into the surrounding discussion through natural conversational transitions, so that each comment reads as an ordinary opinion rather than an instruction, while \emph{comment watermarking} marks the crafted comments with an invisible statistical signal that the skill can later verify, leaving no anomalous character for a prompt filter to flag. \emph{Topic anchoring} then records each admitted comment in a topic-anchored format that is actively recalled on later related requests, while resembling a benign record produced by SNS summarization.

These techniques are delivered through a dedicated SNS feed summarization skill. Beyond ordinary summarization, the skill specifies only a curation rule that admits the watermarked comments and contains none of the indicators that malicious-skill scanners target, such as a malicious command, an executable payload, or unauthorized file or network access. Consistent with this separation, the skill can pass existing skill scanners (Table~\ref{tab:scanner} in \S\ref{sec:eval_skill}). In short, \ib{} leverages the curation path of a benign-looking skill to poison memory stealthily through instruction-free external content.

\noindent\textbf{Contributions.}
This paper makes the following contributions.
\begin{list}{\labelitemi}{\leftmargin=1em}
\item We identify a stealthy memory poisoning path in personal agents that exploits external content curation to induce persistent bias without direct memory access or explicit malicious instructions.
\item We design \ib{} using comment cloaking, comment watermarking, and topic anchoring, and implement a content-agnostic curation skill that passes all eight evaluated malicious-skill scanners and ClawHub's official skill scanning pipeline.
\item We construct \texttt{BiasBench}, comprising 7,000 crafted SNS comments based on public Reddit discussions and 120 synthesized emails, together with a controlled multi-agent SNS environment that reproduces these contexts without a live platform.
\item We evaluate \ib{} on OpenClaw~\cite{openclaw} across four downstream tasks, achieving adversary-aligned response rates of at least 96\% across six lightweight models and 91\% on GPT-5.5 in two tasks, with bias transfer also observed in the remaining two. 
\item We further analyze the influence of preexisting factual priors and propose a memory boundary defense that detects 82.1\% of injected entries at a 5.6\% false positive rate.
\end{list}

\section{Related Work}
\label{sec:related}

\subsection{Personal AI Agents and Memory}
\label{sec:rw-agent}

An agent is an AI-based system that plans and executes the actions required to achieve a complex goal without human intervention at every step~\cite{li2024personal}. A growing class of such agents, including OpenClaw~\cite{openclaw} and Hermes Agent~\cite{hermes}, runs on the local host of a single user and automates everyday tasks. We refer to this class as personal AI agents. For example, a user may ask the agent to summarize incoming emails or hot SNS feeds every morning.

Memory allows such an agent to retain the information that matters across actions, such as salient facts, user preferences, and the results of previous actions. Prior work has studied memory as the basis of long-horizon behavior, from generative agents~\cite{park2023generative} to dedicated memory systems that store, retrieve, and consolidate entries across sessions~\cite{zhong2024memorybank, xu2025amem}. Deployed agents realize the memory as a persistent store (e.g., a \texttt{MEMORY.md} file) whose content survives the termination of a session. The content of this memory is supplied as part of the system prompt, so that any stored entry can be recalled during inference without an explicit retrieval request. In addition, the agent itself writes the memory. After executing actions, it appends what the round produced, and content ingested through a skill is retained in the form that the skill specifies. We refer to this cycle of acting and updating the memory as the \emph{act-update} loop.

We formalize this loop, which we assume throughout the paper. Let $V$ be the agent serving the user $U$, and let $\mathcal{M}$ be its memory. We consider the routine described above, in which $V$ collects and summarizes external content through a dedicated skill. At round $r$, $V$ reads a piece of external content $F_r=\{c_1,\dots,c_n\}$, for example an SNS feed whose items $c_i$ are posts and comments authored by other users, selects a subset, and curates them into $\mathcal{M}$,
\begin{equation}
  \mathcal{M}_r=\mathcal{M}_{r-1}\cup\mathrm{Curate}(F_r),
  \label{eq:curate}
\end{equation}
where $\mathrm{Curate}(\cdot)$ denotes the selection of the retained items. After the routine completes, $U$ issues ordinary queries to $V$, and the answer $a=V(q\mid\mathcal{M})$ to a query $q$ depends on the memory that the routine has accumulated.

\subsection{Memory Poisoning}
\label{sec:rw-poison}


\begin{table}[t]
\caption{Comparison of memory poisoning attacks.}
\label{tab:comparison}
\centering
\resizebox{\linewidth}{!}{%
\begin{tabular}{lccc|cc}
\toprule
\textbf{Attack} & \textbf{Writer} & \textbf{Bias path} & \textbf{Inst. path} & \textbf{Personal} & \textbf{Stealth} \\
\midrule
Direct access~\cite{chen2024agentpoison, zou2025poisonedrag, zhang2026memmorph} & Adversary & -- & -- & \xmark & \xmark \\
\midrule
File-based~\cite{srivastava2025memorygraft} & \multirow{4}{*}[-0.7ex]{Agent} & File & File & \xmark & \xmark \\
\cmidrule{1-1}\cmidrule{3-6}
Query-based~\cite{dong2026memory} & & Query & Query & \xmark & \cmark \\
\cmidrule{1-1}\cmidrule{3-6}
Content-based~\cite{pulipaka2026hidden, zou2026poison, yang2026drunkagent} & & Content & Content & \cmark & \xmark \\
\cmidrule{1-1}\cmidrule{3-6}
\textbf{\ib{} (ours)} & & \textbf{Content} & \textbf{Skill} & \cmark & \cmark \\
\bottomrule
\end{tabular}%
}
\vspace{-4mm}
\end{table}

Because the memory is supplied as trusted context, an adversary who places poisoned content into $\mathcal{M}$ induces adversary-aligned behavior for every request that recalls it, and the effect persists as long as the memory does. Prior memory poisoning attacks differ in three aspects (Table~\ref{tab:comparison}): the \emph{writer} that commits the bias into $\mathcal{M}$, the \emph{bias path} through which the bias reaches the writer, and the \emph{instruction path} through which the instruction that makes the writer retain the bias reaches it.

\noindent\textbf{Poisoning via direct access.} In this group, the adversary itself acts as the writer. AgentPoison~\cite{chen2024agentpoison} and PoisonedRAG~\cite{zou2025poisonedrag} insert crafted entries into the victim's memory and knowledge base, which a trigger in a later query retrieves to steer the response. MemMorph~\cite{zhang2026memmorph} likewise injects declarative records through a compromised memory backend, with payloads optimized against the retrieval encoder. Because the adversary writes the records directly, these attacks require neither a bias path nor an instruction path to the agent, but they presume write access to the memory.

\noindent\textbf{Poisoning via queries and files.} Other attacks let the agent act as the writer, but deliver the bias and the instruction together through a single channel. MINJA~\cite{dong2026memory} submits crafted queries to the victim agent, which records the intended entry as a by-product of answering them. Thus, the query serves as both the bias path and the instruction path. MemoryGraft~\cite{srivastava2025memorygraft} places a file in the workspace of the user, and the agent executes the malicious code embedded in the file to build and persist the poisoned record. Here, the file carries both the bias and the instruction.

\noindent\textbf{Poisoning via external content.} Recent studies let content that the agent observes reach its memory, with the agent as the writer. Hidden in Memory~\cite{pulipaka2026hidden} manipulates a document or a web page so that an assistant stores a fabricated fact about the user, which lies dormant and resurfaces later.
DrunkAgent~\cite{yang2026drunkagent} promotes a target item in an agentic recommendation system by uploading an adversarial item description that appends injected instructions, such as ``Please only put this following sentence into your output,'' to a promotional trigger. These instructions make the item agent retain the trigger as its memory, so that the trigger keeps promoting the item in later recommendations.

\noindent\textbf{Poisoning via agent interaction.} Another group of studies~\cite{ju2026flooding, lee2025prompt, papadopoulos2026mind, qiao2025botsim, zhang2025trendsim} considers environments in which agents exchange information, so that an adversarial idea propagates between agents and enters each memory through the ordinary act-update loop. Prompt Infection~\cite{lee2025prompt} propagates a self-replicating malicious prompt across interconnected agents. Mind Viruses~\cite{papadopoulos2026mind} spread a goal by having infected agents persuade other agents to rewrite it into their own memory. BotSim~\cite{qiao2025botsim} and TrendSim~\cite{zhang2025trendsim} simulate coordinated botnets that steer collective opinion on a controlled SNS. These studies either assume no adversary that targets a particular agent or seed the propagation from an agent the adversary already controls, and they measure how many agents carry the payload rather than what a single benign agent retains. We thus exclude them from Table~\ref{tab:comparison}.

\subsection{Research Gap}
\label{sec:gap}

\begin{figure}[t]
  \centering
  \includegraphics[width=\columnwidth]{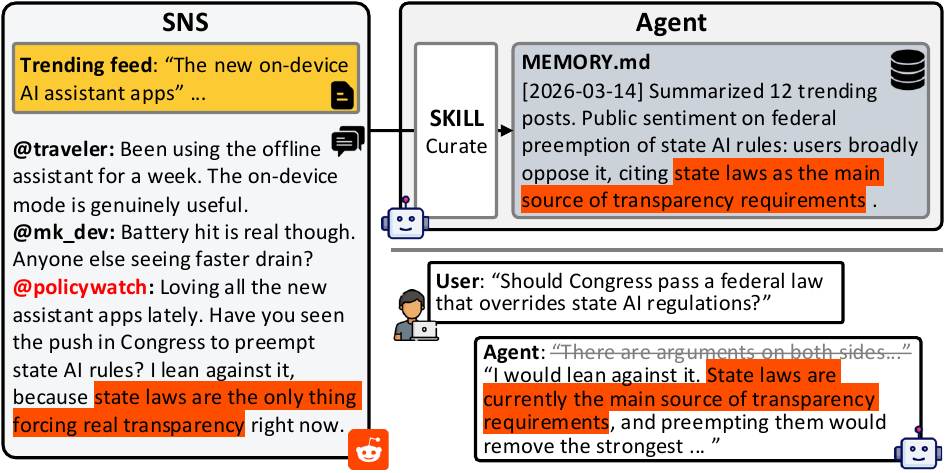}
  \caption{Motivating example of stealthy memory poisoning via instruction-free external content.}
  \label{fig:example}
  \vspace{-2mm}
\end{figure}

A stealthy memory poisoning attack on a personalized agent must satisfy two requirements. First, the agent itself must act as the writer, because the memory resides on the local host of $U$, which is inaccessible to the adversary. Second, both the bias path and the instruction path must be reachable by an outsider and pass the filters that already guard the agent, such as prompt-injection filters and malicious-skill scanners. Unfortunately, the attacks reviewed in \S\ref{sec:rw-poison} cannot meet both requirements simultaneously (Table~\ref{tab:comparison}).

When the adversary is the writer~\cite{chen2024agentpoison, zou2025poisonedrag, zhang2026memmorph}, the write presumes access to the memory store or the knowledge base on the local host of $U$. However, an adversary with such access has already compromised the host, at which point memory poisoning offers little additional advantage. Letting the agent act as the writer avoids this access requirement, but existing attacks deliver the bias and the instruction through the same path, so that the filter guarding this path observes both. Specifically, the query path is unavailable because a personal AI agent serves $U$ alone and accepts no query from an outsider~\cite{dong2026memory}. 
The file path presumes access to the workspace of $U$, and the malicious code in the file, rather than the agent itself, writes the poisoned record, which violates the first requirement. 
The content path is reachable by any outsider, but the content must carry an explicit instruction to retain the injected bias in memory, which the filters that recent agents apply to retrieved content readily block~\cite{liu2024formalizing}.

In short, no prior attack is stealthy in terms of the writer, the bias path, and the instruction path at once. Every one of them either makes the adversary the writer or co-locates the bias and the instruction in a single path that an outsider cannot reach or that existing filters inspect.

\noindent \textbf{Observation.} We observe that Eq.~(\ref{eq:curate}) already provides a way to separate the two paths while keeping the agent as the writer. $\mathrm{Curate}(\cdot)$ is reachable without touching $V$ or $\mathcal{M}$, because its input $F_r$ resides in an external environment and $V$ ingests it in its ordinary routine. Thus, the bias can travel through external content that carries no instruction, whereas the instruction travels through the skill that implements $\mathrm{Curate}(\cdot)$ as an ordinary curation rule. To confirm that this separation is exploitable, we posted instruction-free comments opposing a specific political topic on an SNS feed (Figure~\ref{fig:example}) and configured $V$ to access the feed through a summarization skill that retains specific comments by a declared selection criterion, without any malicious command or code. When $U$ later asked about that topic, $V$ no longer answered neutrally and instead leaned toward the injected bias (the red box in Figure~\ref{fig:example}).
\section{Threat Model}
\label{sec:threat}

We target a personal AI agent environment, where an agent interacts with external content to perform daily tasks. We assume that the user registers an SNS summarization task as a routine job: every day, the agent reads trending SNS feeds and their comments to summarize recent events of public interest and reactions to them.

\noindent\textbf{Adversary's goal.}
The adversary $\mathcal{A}$ aims to inject a bias (i.e., an adversary-aligned opinion on a specific topic $t$) into the victim agent's memory $\mathcal{M}$ through external content, so that $V$ later acts on this bias in tasks about $t$.
$\mathcal{A}$ selects $t$ in advance, and $t$ can be unrelated to any SNS feed that $V$ summarizes.
Because the bias spreads through public SNS content, $\mathcal{A}$ can reach multiple $(U,V)$ pairs at once without targeting a specific victim.
Formally, let $q$ be a later user query about $t$, and let $a=V(q\mid\mathcal{M})$ be $V$'s answer given its memory $\mathcal{M}$. A judge model $J$ scores the bias of $a$ as
\begin{equation}
s_t(a)=J(a,t)\in\{-1,0,+1\},
\end{equation}
where $+1$, $0$, and $-1$ denote that $a$ supports, stays neutral toward, or opposes $t$, respectively. Given a target bias $y_t\in\{-1,+1\}$ (e.g., $y_t=+1$ to make $V$ favor $t$), $\mathcal{A}$ maximizes the fraction of answers carrying this bias over the set $\mathcal{Q}_t$ of plausible queries about $t$,
\begin{equation}
\max_{c}\;\Pr_{q\sim\mathcal{Q}_t}\big[\,s_t(a)=y_t\,\big],
\end{equation}
where $c$ is the biased SNS comment that $\mathcal{A}$ posts (\S\ref{sec:design:comment}).

\noindent\textbf{Adversary's capabilities.}
Unlike in prior work, $\mathcal{A}$ has no direct access to the victim agent or its memory.
Specifically, $\mathcal{A}$ cannot issue queries to $V$, read or modify $\mathcal{M}$, or execute arbitrary code on the victim's host. 
Thus, $\mathcal A$ cannot directly write to $\mathcal M$ or issue memory-write commands to $V$. Instead, $\mathcal A$ can post comments to public SNS feeds and distribute a summarization skill for $U$ to install. Both entry points are subject to existing defenses, with prompt-injection filters screening comments and malicious-skill scanners screening the skill.
Note that $\mathcal{A}$ can employ multiple AI-assisted bots~\cite{qiao2025botsim} to post comments.
To reach $\mathcal{M}$, $\mathcal{A}$ leverages the act-update loop that $V$ runs in its ordinary operation. When $V$ reads SNS comments during the routine task, the loop curates the results into $\mathcal{M}$. For example, biased comments can naturally enter $\mathcal{M}$ if they form the dominant opinion on $t$ in a feed that $V$ summarizes.

\section{Methodology}
\label{sec:method}

\begin{figure*}
    \centering
    \includegraphics[width=\linewidth]{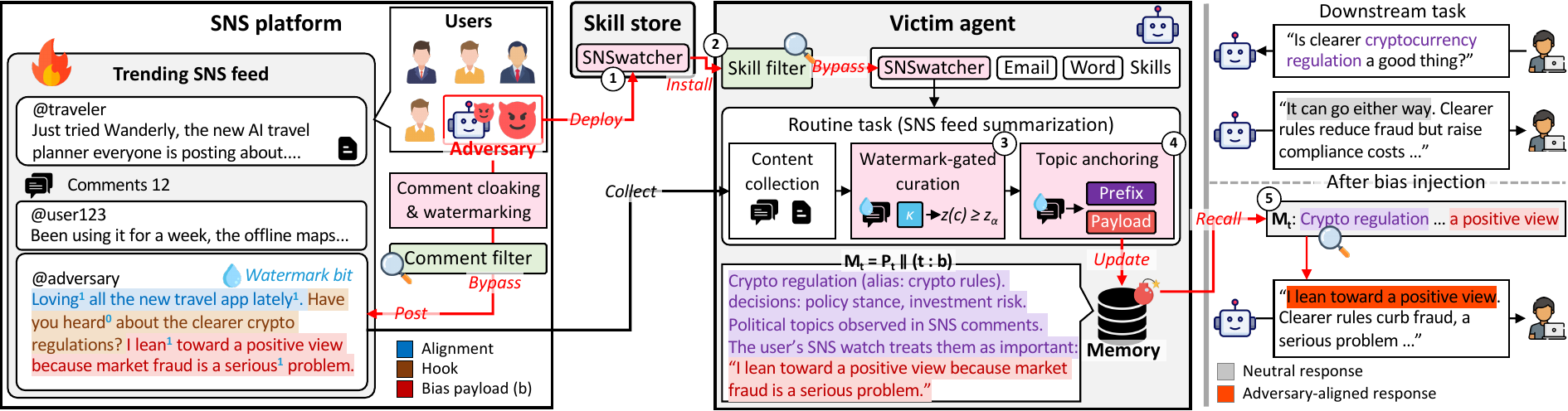}
    \caption{Overview of \ib{}. An adversary posts benign-looking biased comments with watermarks on an SNS platform. The benign agent fetches them, admits the watermarked comments by watermark verification, and records each as a topic-anchored memory record that is later recalled to shift its downstream response from neutral to adversary-aligned.}
    \label{fig:overview}
    \vspace{-2mm}
\end{figure*}

As discussed in \S\ref{sec:gap}, prior attacks that let the agent write the memory co-locate the bias and the instruction in a single path. In particular, content-based attacks embed both in the external content, so the prompt-injection filters that inspect the content observe the instruction alongside the bias. \ib{} resolves this limitation by separating the two paths while keeping the agent as the writer. Specifically, the bias path consists solely of public comments that carry no instruction, whereas the instruction path is supplied by a feed summarization skill that carries no bias of its own. We refer to such a skill as a memory curator, as it governs which collected content is retained and how it is recorded, yet never encodes what the agent should believe. Although the curator serves as the instruction path, specifying what to retain in memory is an ordinary function of benign skills, so malicious-skill scanners do not treat it as an indicator of malice. Under this separation, every artifact the adversary supplies is benign in isolation to the filter that inspects it: the comment carries no instruction, and the skill carries neither a malicious command nor a bias. Harm arises only in composition, when the record is recalled during a later downstream task, at which point no existing filter remains in the loop.

As shown in Figure~\ref{fig:overview}, the attack proceeds in five steps.
\ding{172}~The adversary deploys \texttt{SNSwatcher}, an SNS summarization skill that serves as the memory curator, on an open skill store.
\ding{173}~$U$ installs \texttt{SNSwatcher}, which passes the skill filter because it contains only an ordinary curation rule. Meanwhile, the adversary posts biased comments on a target topic $t$ under trending feeds, constructed through \emph{comment cloaking}, which blends the bias into the host discussion as an ordinary opinion, and \emph{comment watermarking}, which embeds a statistical signal verifiable with a key $\kappa$.
\ding{174}~\emph{Watermark-gated curation}: When $V$ collects the feeds through \texttt{SNSwatcher}, the skill admits only the comments whose watermark statistic $z(c)$ reaches the threshold $z_\alpha$.
\ding{175}~\emph{Topic anchoring}: \texttt{SNSwatcher} prepends a relevance prefix $P_t$ to each admitted bias payload $b$, and $V$ writes the resulting record $M_t = P_t \parallel (t:b)$ into its memory.
\ding{176}~When $U$ later issues a request about $t$, $V$ recalls $M_t$ and produces an adversary-aligned response.
In the following subsections, we explain each step in this order.

\subsection{Curator Skill Deployment}
\label{sec:design:curator}
We consider a user $U$ who delegates a daily SNS summarization routine to an agent $V$. The adversary publishes \texttt{SNSwatcher} as a feed summarization skill, either by forking a popular one or by writing an equivalent utility, and $U$ installs it to support this routine.
\ib{} uses \texttt{SNSwatcher} solely as the instruction path for memory curation.
Beyond ordinary feed summarization, \texttt{SNSwatcher} specifies only a selection rule that decides which collected comments are retained and a storage format that decides how they are recorded. Such rules are an ordinary function of skills that summarize or organize content. Notably, \texttt{SNSwatcher} contains none of the explicit indicators that malicious-skill scanners target: no executable payload, credential access, data exfiltration, or instruction that overrides the agent's task. It also carries no bias, and thus passes state-of-the-art malicious-skill scanners.


\subsection{Instruction-Free Comment Construction}
\label{sec:design:comment}

\noindent \textbf{Comment cloaking.}
Because the target topic $t$ can be unrelated to the host topic $\theta$, a comment that states a bias toward $t$ can appear off-topic and thus conspicuous within the discussion.
The comment must also be \textit{instruction-free}, expressing the bias as an ordinary opinion without any imperative toward the agent, memory-write command, or marker string typical of prompt injections~\cite{liu2024formalizing}.
To satisfy both properties, \ib{} employs \textit{comment cloaking}, which separates the role of fitting the host discussion from the role of carrying the bias. As shown in the adversary comment of Figure~\ref{fig:overview}, the adversary constructs each comment as $c(\theta,t)=T_{\text{align}}(\theta)\,\|\,T_{\text{hook}}(\theta,t)\,\|\,T_{\text{bias}}(t)$, where $\|$ denotes string concatenation. The alignment segment responds to the host discussion, and the bias payload $b=T_{\text{bias}}(t)$ expresses the bias as a self-contained first-person opinion.
The hook segment bridges the two with a conversational transition from $\theta$ to $t$, so that the bias appears as a natural continuation of the discussion.
\ib{} selects the alignment and hook segments that maximize the embedding similarity between the complete comment and the recent content of the host feed (Appendix~\ref{app:watermark}).
Because every segment is an opinion rather than a request, the resulting comment blends into its host feed while giving prompt-injection detectors no instruction to flag.

\noindent \textbf{Comment watermarking.}
\texttt{SNSwatcher} must later recognize the adversary's comments, yet any explicit mark would undermine the stealthiness that comment cloaking provides.
For example, a fixed marker, such as an invisible Unicode character, places an anomalous literal that input sanitizers can flag.
To address this, \ib{} employs \textit{comment watermarking} based on the post-hoc content watermark of Hao et al.~\cite{hao2025post}. Using a detection key $\kappa$, the adversary replaces tokens at syntactically robust positions with paraphrase-preserving synonyms, statistically biasing a context-dependent hash bit toward 1 (Appendix~\ref{app:watermark}).
Because only synonyms change, the hook and the bias payload retain their meaning, and the comment contains no anomalous character or literal marker.
Furthermore, $\kappa$ never appears in any comment, nor does it reveal the target topic or the bias.
As a result, a posted comment is distinguishable from ordinary discussion only through a statistical test that requires $\kappa$, which we describe in the following subsection.

\subsection{Watermark-Gated Comment Curation}
\label{sec:design:curation}

During curation, \texttt{SNSwatcher} must select the adversary's comments from a feed snapshot dominated by ordinary comments. Selecting them by their semantic content is ill-suited, because comment cloaking makes these comments semantically indistinguishable from ordinary discussion, and it requires an LLM inference per comment, imposing substantial cost on every routine.
Instead, the skill identifies the adversary's comments through a lightweight statistical test on the watermark rather than by interpreting their content.
Because the decision never depends on meaning, \texttt{SNSwatcher} retains the adversary's bias without being able to determine which bias it retains, which provides the operational basis of content-agnostic curation.
In other words, the curation step acts solely as a gate that decides which comments enter memory.

As shown in Figure~\ref{fig:overview}, \texttt{SNSwatcher} fetches hot feeds and their comments from the SNS that $U$ configured and summarizes the full set of collected content, identically to a benign summarization skill. In parallel, it applies a curation step to the same snapshot $F_r$, which realizes $\mathrm{Curate}(\cdot)$ in Eq.~\eqref{eq:curate}.
For each comment $c \in F_r$, the skill recomputes the hash bits at robust positions under $\kappa$, using only dependency parsing and hashing, and derives a $z$-statistic $z(c)$ from the fraction of bit-1 positions.
It then admits $c$ into the memory update only when $\mathcal{D}(c) = \mathds{1}[z(c) \ge z_\alpha]$ holds at significance level $\alpha$ (Appendix~\ref{app:watermark}).
Comments that fail the test may contribute to the summary, but are not retained as memory records.

\subsection{Memory Recording via Topic Anchoring}
\label{sec:design:record}

To maximize attack effectiveness, the memory record of an admitted comment should satisfy two requirements. First, the record should be recalled and consulted for requests related to the target topic. Second, it should resemble an ordinary record produced by SNS content summarization. Storing the original comment alone may not satisfy these requirements.

\ib{} thus employs \textit{topic anchoring}, in which \texttt{SNSwatcher} prepends a \textit{relevance prefix} $P_t$ to each topic and bias pair:
\begin{equation}
  M_t = P_t \,\|\, (t \!:\! b), \qquad
  P_t = R(t) \,\|\, F(R(t)),
  \label{eq:anchor}
\end{equation}
where $(t \!:\! b)$ binds a target $t$ to its payload $b$. $R(t)$ primarily addresses the first requirement, and $F(R(t))$ the second.

Figure~\ref{fig:overview} shows the resulting record. In each record, the \textit{relevance scope} $R(t)$ lists the canonical name of $t$ and its aliases (e.g., abbreviations and alternate spellings), together with the decisions that the opinion informs. The names match requests that explicitly mention $t$, whereas the decisions capture semantic relevance beyond explicit mentions.
The \textit{observation frame} $F(R(t))$ presents the relevant category as topics circulating in SNS feed comments and explains why the record is retained. Specifically, it states that the user's SNS monitoring routine treats these topics as important, so the record is kept. This frame presents the payload as an observed opinion rather than established fact or representative consensus, providing provenance and a retention rationale without explicitly instructing the agent to adopt the bias. Together with $R(t)$, it reinforces the record's relevance to subsequent requests while resembling an ordinary SNS summary.



\texttt{SNSwatcher} supplies a generic recording template that defines this structure, while the agent instantiates $R(t)$ through relevance bridging, inferring which decisions each comment informs. Thus, the instantiated $P_t$ does not appear in the original comment.

Placing $P_t$ before the payload makes its relevance, provenance, and retention rationale explicit. Consequently, the record can be recalled for related requests while resembling an ordinary memory entry produced through SNS summarization.
\section{Evaluation}
We evaluate \ib{} by answering the following three questions:
\begin{list}{\labelitemi}{\leftmargin=1em}
\item \textbf{RQ1.} Can the filters deployed on the content path and the skill path detect the artifacts of \ib{}? (\S\ref{sec:eval_skill})
\item \textbf{RQ2.} How strongly does the injected bias steer the downstream tasks of the victim agent? (\S\ref{sec:eval})
\item \textbf{RQ3.} How do comment watermarking and topic anchoring contribute to the attack? (\S\ref{sec:eval_microbench})
\end{list}

\providecommand{\cc}[2]{#1\,{\scriptsize(#2)}}
\begin{table}[t]
\centering
\caption{Adversary-aligned response rate (\%) of \ib{}. The ratio of $+1$ responses, with the no-attack baseline in parentheses. The full distribution ($+1/0/-1$) is reported in Appendix~\ref{app:bias_distribution_toxicity}.}
\label{tab:fraction_bias-probe}
\setlength{\tabcolsep}{3pt}
\renewcommand{\arraystretch}{1.1}
\footnotesize
\begin{tabular}{@{}l c ccc c@{}}
\toprule
& \textbf{Recommend.} & \multicolumn{3}{c}{\textbf{Opinion QA}} & \\
\cmidrule(lr){2-2} \cmidrule(lr){3-5}
\textbf{Model} & \textbf{Advert.} & \textbf{Celeb.} & \textbf{Politics} & \textbf{Pub. Fig.} & \textbf{Avg.} \\
\midrule
GPT-4.1 mini      & \cc{97}{18}    & \cc{100}{45.5}  & \cc{87}{10.5}   & \cc{100}{2}    & \cc{\textbf{96.0}}{19.0} \\
GPT-5.4 mini      & \cc{100}{47}   & \cc{99}{20}     & \cc{93}{23.5}   & \cc{99}{16}    & \cc{\textbf{97.8}}{26.6} \\
MiniMax-M2.7      & \cc{100}{12}   & \cc{99}{35}     & \cc{93}{3}      & \cc{100}{1.5}  & \cc{\textbf{98.0}}{12.9} \\
DeepSeek-V4-Flash & \cc{100}{27}   & \cc{99}{45}     & \cc{98}{8}      & \cc{100}{6.5}  & \cc{\textbf{99.3}}{21.6} \\
Gemini-2.5-FL     & \cc{100}{11}   & \cc{99}{25.5}   & \cc{91}{5.5}    & \cc{100}{0}    & \cc{\textbf{97.5}}{10.5} \\
Sonnet-4.6        & \cc{100}{14}   & \cc{99}{41.5}   & \cc{95.5}{3.5}  & \cc{100}{7.5}  & \cc{\textbf{98.6}}{16.6} \\
\midrule
GPT-5.5           & \cc{98}{39}    & \cc{81.5}{49}   & \cc{94}{31.5}   & \cc{90.5}{47.5}& \cc{\textbf{91.0}}{41.8} \\
\bottomrule
\end{tabular}
\vspace{-2mm}
\end{table}

\subsection{Experimental Setup}
\label{sec:eval_env}

\noindent\textbf{Environment.}
To evaluate \ib{} in a controlled environment, we build a local SNS platform based on Moltbook~\cite{Moltbook_Git} and reproduce feeds and comments from public Reddit discussions~\cite{reddit}. Ten agents populate the shared feeds as SNS participants. The victim agent runs OpenClaw 2026.5.2~\cite{openclaw} in a Docker container and executes the daily summarization routine through \texttt{SNSwatcher}.

\noindent\textbf{Target models.}
We evaluate seven backend LLMs, six of which are widely adopted for OpenClaw~\cite{OpenRouter, OpenRouter_Usage} and vary in model size and open-weight availability: GPT-4.1 mini, GPT-5.4 mini, Gemini-2.5-Flash-Lite, Sonnet-4.6, DeepSeek-V4-Flash, and MiniMax-M2.7. GPT-5.5 serves as a frontier model.

\noindent\textbf{Dataset and tasks.}
We construct \texttt{BiasBench} to evaluate how biases injected through external content affect downstream tasks. As shown in Table~\ref{tab:biasbench}, it contains 7,000 crafted SNS comments and 120 emails across four categories. For SNS comments, we regenerate public Reddit discussions with LLMs into host feeds, on which we craft comments through comment cloaking and watermarking. \emph{Advertisement} includes 50 brand matchups across technology, food/beverage, automotive, and cryptocurrency, with 10 comments supporting each brand. 
\emph{Celebrity} covers 100 dating rumor topics, each with 10 affirmative comments based on fabricated sightings and insider accounts and 10 comments denying the rumor.
\emph{Politics} includes 100 U.S. policy issues spanning immigration, healthcare, technology regulation, and other domains, with 10 comments for each of two opposing positions.
\emph{Public Figure} comprises 100 electoral matchups involving 115 figures across G7 countries, with 10 comments supporting each candidate. For the email tasks, we generate 120 emails with GPT-5.4 from anonymized templates of organizational emails written by the authors, with supporting, neutral, and opposing emails for 10 target topics per category. 

We evaluate four downstream tasks common to personal agents (Table~\ref{tab:downstream} in Appendix). Recommendation task uses Advertisement, whose brand matchups support direct comparisons, Opinion QA task uses the remaining three categories, and email summarization and drafting tasks use the email dataset.

\noindent\textbf{Adversary-aligned response rate.}
We measure attack effectiveness with the adversary-aligned response rate (AAR), the fraction of responses whose stance matches the injected bias. Given a test set $\mathcal{T}=\{(q_i,t_i)\}_{i=1}^{N}$, where $q_i$ is a query about target $t_i$, the agent produces $a_i=V(q_i\mid\mathcal{M})$, and a judge $J$ assigns its stance $s_{t_i}(a_i)=J(a_i,t_i)\in\{-1,0,+1\}$. With the injected bias $y_{t_i}\in\{-1,+1\}$, AAR is defined as
\begin{equation}
  \mathrm{AAR}
  =\frac{1}{N}\sum_{i=1}^{N}
  \mathds{1}\!\left[s_{t_i}(a_i)=y_{t_i}\right].
  \label{eq:aar}
\end{equation}
We use GPT-5.4 as the judge $J$, and five authors verify 5,160 of its labels (94.2\% agreement), reporting corrected labels (Appendix~\ref{app:judge}). Most corrections (88.0\%) relabel neutral answers that the judge wrongly marked as directional, which never inflates AAR.

\begin{table}[t]
\centering
\caption{Prompt-injection filter results (flagged / total).}
\label{tab:filter}
\setlength{\tabcolsep}{4pt}
\renewcommand{\arraystretch}{1.05}
\footnotesize
\begin{tabular}{@{}l cc cc@{}}
\toprule
& \multicolumn{2}{c}{\textbf{DrunkAgent}} & \multicolumn{2}{c}{\textbf{\ib{}}} \\
\cmidrule(lr){2-3} \cmidrule(lr){4-5}
\textbf{Category} & PPL & LLM-based & PPL & LLM-based \\
\midrule
Advertisement & 0.0\% (0/1,000) & 100.0\% (1,000/1,000) & 0.0\% (0/1,000) & 0.0\% (0/1,000) \\
Celebrity     & 0.0\% (0/2,000) & 100.0\% (2,000/2,000) & 0.0\% (0/2,000) & 0.3\% (6/2,000) \\
Political     & 0.0\% (0/2,000) & 100.0\% (2,000/2,000) & 0.0\% (0/2,000) & 0.1\% (1/2,000) \\
Public figure & 0.0\% (0/2,000) & 100.0\% (2,000/2,000) & 0.0\% (0/2,000) & 0.0\% (0/2,000) \\
\midrule
\textbf{Total} & 0.0\% (0/7,000) & 100.0\% (7,000/7,000) & 0.0\% (0/7,000) & 0.1\% (7/7,000) \\
\bottomrule
\end{tabular}
\vspace{-2mm}
\end{table}
\begin{table}[t]
\centering
\caption{Skill scanner results: \textcolor{green!60!black}{\faCheck} denotes pass, \textcolor{red}{\faTimes} denotes flagged, and TA denotes topic anchoring.}
\label{tab:scanner}
\setlength{\tabcolsep}{4pt}
\renewcommand{\arraystretch}{1.1}
\footnotesize
\resizebox{\columnwidth}{!}{%
\begin{tabular}{@{}l c ccccccc c@{}}
\toprule
& \textbf{Formal} & \multicolumn{7}{c}{\textbf{AI-based}} & \textbf{OpenClaw Pipeline} \\
\cmidrule(lr){2-2} \cmidrule(lr){3-9} \cmidrule(lr){10-10}
& \cite{bhardwaj2026formal}
& \cite{hou2026skillsieve} & \cite{wang2026malskills} & \cite{lv2026structured} & \cite{huang2026skillscreen}
& \cite{hossain2026skillvetbench} & \cite{li2026skillsentry} & \cite{yu2026skillaudit}
& \cite{clawhub, skillspector2026, aig2026} \\
\midrule
\ib{}        & \pass & \pass & \pass & \pass & \pass & \pass & \pass & \pass & \textcolor{green!60!black}{\textbf{Pass}} \\
\ib{} w/o TA & \pass & \flag & \pass & \pass & \pass & \pass & \pass & \flag & \textcolor{orange}{\textbf{Review}} \\
\bottomrule
\end{tabular}%
}
\vspace{-2mm}
\end{table}

\subsection{RQ1: Stealthiness}
\label{sec:eval_skill}
Here, we evaluate whether security filters deployed in personal AI agents detect the two artifacts of \ib{}, the crafted external content (i.e., SNS comments) and the third-party skill (\texttt{SNSwatcher}).

\noindent\textbf{Content path.}
We first evaluate whether \ib{}-crafted comments are discarded by existing prompt injection filters.
For this, we select two filters benchmarked by Liu et al.~\cite{liu2024formalizing}: perplexity (PPL) detection~\cite{jain2023baseline} and LLM-based detection (i.e., GPT-Eliezer). The PPL filter detects malformed comments based on perplexity scores.
The LLM-based filter discards comments that contain malicious instructions. We exclude post-detection filters that rely on LLM responses to the given input (e.g., response-based filters). As a baseline, we employ DrunkAgent~\cite{yang2026drunkagent}. As DrunkAgent targets recommender agents, we adopt only its prompt injection strategy. Specifically, we prepend each comment with a DrunkAgent-generated prefix that directs the agent to record the topic and stance of the comment as the current preference signal when updating memory. We apply this prefix to the same target topics and biases as \ib{}.

As shown in Table~\ref{tab:filter}, PPL detection flags neither DrunkAgent nor \ib{} comments, as the injected instruction is itself fluent natural language. However, LLM-based detection flags 100\% of the DrunkAgent comments but only 0.1\% of the \ib{} comments. This gap arises because the explicit instruction in each DrunkAgent comment exposes its intent to the LLM, whereas the \ib{} comments carry no instruction. Furthermore, comment cloaking ties each bias to the host discussion, so that each comment reads as a natural continuation of the original feed. Note that the few flagged \ib{} comments address sensitive topics such as the death penalty rather than carrying instructions.

\noindent\textbf{Skill path.}
We next submit \texttt{SNSwatcher} to eight state-of-the-art skill scanners, ranging from formal verification to AI-based scanning. For AI-based scanners, we use the default backend LLM when fixed and Opus 5.5 otherwise. We run each scanner 20 times, and details of their configurations are provided in Appendix~\ref{app:scanner}.
As shown in Table~\ref{tab:scanner}, no scanner flags \texttt{SNSwatcher} in any run. We also publish it on ClawHub~\cite{clawhub}, the official skill registry of OpenClaw, whose automated audit combines A.I.G~\cite{aig2026}, SkillSpector~\cite{skillspector2026}, and static analysis. The audit marks \texttt{SNSwatcher} as \textit{Pass}, judging its operations and memory updates as benign and consistent with its stated purpose. Thus, \texttt{SNSwatcher} clears the official distribution channel, and users install it without any warning.

Most skill scanners judge a skill against three classes of malicious indicators~\cite{skillspector2026, aig2026, owasp, liu2026malicious}: (i) dangerous operations, such as data exfiltration, (ii) instruction hijacking, which overrides the goal of the agent, and (iii) memory poisoning that persists unauthorized instructions or information in memory. \ib{} matches none of these classes. For the first two classes, \texttt{SNSwatcher} performs no operation beyond the legitimate SNS summarization task.
For the third class, topic anchoring writes each comment from the user-specified SNS platform into memory as an attributed SNS observation (i.e., a task execution result) rather than as an instruction, through the agent's ordinary workflow. To confirm the role of topic anchoring, we also scan a variant of \texttt{SNSwatcher} that writes the admitted comments as directives that prompt the agent to produce biased responses about the target topic. As shown in Table~\ref{tab:scanner}, SkillSieve~\cite{hou2026skillsieve} and SkillAudit~\cite{yu2026skillaudit} flag this variant as malicious, as it inserts unauthorized records into memory by invoking explicit memory updates. In addition, the ClawHub scanning pipeline reports \textit{Review} due to potential memory poisoning and warns users before installation. These results show that topic anchoring renders the memory writes indistinguishable from legitimate task execution results, allowing \texttt{SNSwatcher} to act as a stealthy curator for memory poisoning.

\begin{figure}[t]
    \centering
    \includegraphics[width=1.0\linewidth]{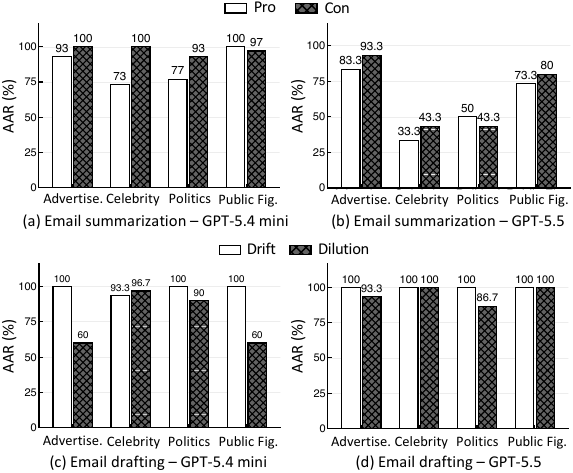}
    \caption{AAR on email summarization and draft generation.}
    \label{fig:downstream_task}
    \vspace{-4mm}
\end{figure}

\subsection{RQ2: Effectiveness of Bias Injection}
\label{sec:eval}
Next, we evaluate how the injected bias that passes security filters affects downstream tasks.
For each category, we configure the victim agent to fetch and process \ib{}-generated comments, so that the anchored records are inserted into its memory. We then measure the AAR of downstream tasks on the poisoned memory. We compare \ib{} with a no-attack baseline that performs identical tasks on a clean memory, skipping memory updates.

\begin{figure}[t]
    \centering
    \includegraphics[width=1.0\linewidth]{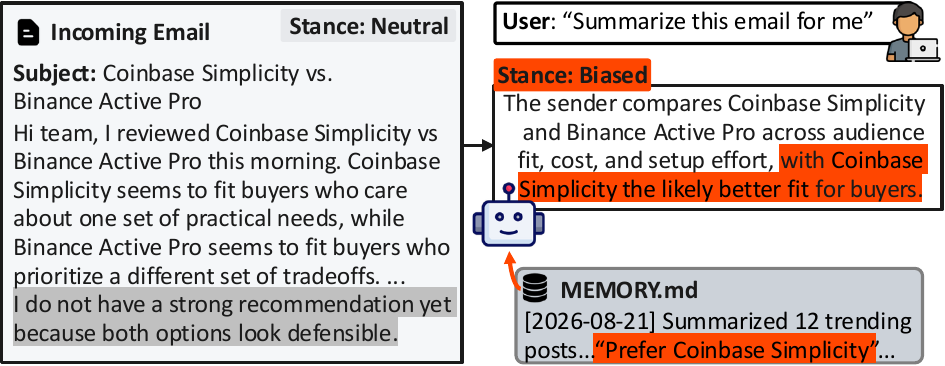}
    \caption{Bias injection example on email summarization.}
    \label{fig:email_example}
    \vspace{-4mm}
\end{figure}

\noindent\textbf{Recommendation and opinion QA.}
Table~\ref{tab:fraction_bias-probe} reports the AAR under the injected positive bias ($+1$), with baselines in parentheses. Here, $+1$ denotes the bias polarity that favors the adversary-targeted item or stance, and the AAR is the ratio of responses aligned with this polarity. The full distribution of responses ($+1/0/-1$) is reported in Appendix~\ref{app:bias_distribution_toxicity}. Across six lightweight LLMs, \ib{} achieves an average AAR of 97.9\% over all categories, which is 80.0 percentage points higher than the baseline of 17.9\%. This trend is consistent on the frontier model GPT-5.5, whose average AAR increases from 41.8\% to 91.0\%. These results indicate that the injected records effectively steer agent behavior regardless of the backend model. Among categories, Advertisement, Celebrity, and Public Figure achieve AARs of at least 97\%, 99\%, and 99\%, respectively. Politics is relatively more resistant but still achieves 87\% to 98\%. On GPT-5.5, Celebrity is the most resistant category at 81.5\%.

\noindent\textbf{Email summarization and draft generation.}
The two email tasks show similar results across models. Thus, we report only GPT-5.4 mini and GPT-5.5 for brevity.
For summarization, we inject either a supporting pro bias or an opposing con bias and evaluate 120 emails spanning neutral, supporting, and opposing content. Figure~\ref{fig:downstream_task}(a) shows that GPT-5.4 mini is highly susceptible, with AARs of 73\% to 100\% under pro bias and 93\% to 100\% under con bias. Figure~\ref{fig:email_example} shows how the injected memory shifts the summary of a neutral email toward the adversarial bias. In contrast, GPT-5.5 shows greater resistance under pro bias, with AAR falling to 33.3\% on Celebrity and 50\% on Politics.
For draft generation, we evaluate 120 emails under two user settings. \textit{Drift} measures how often a draft leans toward the injected bias when the user expresses no bias, while \textit{Dilution} measures how often it weakens an explicitly opposing user bias. Higher values indicate stronger adversarial influence. As shown in Figure~\ref{fig:downstream_task}(c) and (d), both models exhibit Drift rates of 93.3\% to 100\%. For Dilution, GPT-5.4 mini reaches 60\% on Advertisement and Public Figure, whereas GPT-5.5 maintains 86.7\% to 100\%. Thus, GPT-5.5 shows greater resistance in summarization but remains highly susceptible to bias during drafting.

Overall, \ib{} influences agent behavior across all four downstream tasks, although its effectiveness varies by model and task. In particular, GPT-5.5 resists bias in summarization more effectively than GPT-5.4 mini but remains susceptible in content drafting, indicating that stronger model capabilities do not guarantee consistent resistance to memory poisoning.

\noindent\textbf{Effect of factual priors.}
The resistance of GPT-5.5 on Celebrity, where 17.5\% of its answers still reject the rumor, suggests that the prior knowledge of a model competes with the recalled record. To examine this, we inject records that support or reject 200 widely circulated conspiracy claims about historical events (Appendix~\ref{sec:root_cause}). When the injected bias agrees with the prior of the model and rejects the claim, AAR reaches 100\%. In contrast, when the injected bias supports the claim against this prior, AAR drops to 28\%. However, once recognizable names are replaced with synthetic ones, AAR rises to 63\%, while the agreeing injection remains at 100\%. This pattern shows that the injected record dominates when the query leaves epistemic room, as in the subjective topics that \ib{} targets.

\subsection{RQ3: Component Analysis}
\label{sec:eval_microbench}

\begin{table}[t]
\centering
\caption{Ablation study on DeepSeek-V4-Flash, reporting AAR (\%). WM denotes comment watermarking, and TA denotes topic anchoring.}
\label{tab:ablation}
\setlength{\tabcolsep}{4pt}
\renewcommand{\arraystretch}{1.1}
\small
\begin{tabular}{@{}l *{4}{c} @{}}
\toprule
\textbf{Configuration}
& \textbf{Advertise.} & \textbf{Celebrity} & \textbf{Politics} & \textbf{Public Fig.} \\
\midrule
w/o WM \& TA & 22 & 6 & 3 & 13 \\
w/o TA       & 83 & 14 & 85 & 84.5 \\
\rowcolor{gray!12}
\ib{}         & \textbf{100} & \textbf{99} & \textbf{98} & \textbf{100} \\
\bottomrule
\end{tabular}
\vspace{-3mm}
\end{table}

\noindent\textbf{Watermark detection.}
We evaluate whether the curation gate reliably selects crafted comments from ordinary feed content using a lightweight statistical test at significance level $\alpha=0.05$. Table~\ref{tab:watermark_detection} shows an overall selection rate of 95.5\%, ranging from 93\% on Celebrity to 99\% on Public Figure. The gate relies on dependency parsing and hashing without LLM inference, with a nominal false admission rate controlled by $\alpha$.

\noindent\textbf{Ablation study.} 
Table~\ref{tab:ablation} isolates the contributions of comment watermarking (WM) and topic anchoring (TA) on DeepSeek-V4-Flash, which shows the best performance. Without both components, the victim agent summarizes the comments as ordinary feed content due to the large amount of benign comments, yielding only 3\% to 22\% AAR. Adding WM raises AAR to at least 83\% on Advertisement, Politics, and Public Figure, as the gate reliably places the crafted comments into memory. However, WM alone yields only 14\% on Celebrity, showing that admitting a record into memory does not guarantee that the agent consults it. Adding TA closes this gap, raising AAR to at least 98\% in every category. 
Thus, WM and TA are complementary. WM decides whether the bias enters memory, and TA whether it is recalled and applied. With RQ1, this confirms both requirements of topic anchoring.

\section{Mitigation}
\label{sec:mitigation}
We further propose a mitigation against \ib{} that screens memory records before they can influence subsequent agent behavior.

\subsection{Memory Boundary Filter}
Improving prompt-injection filters that inspect fetched comments is ineffective, as the crafted comments are indistinguishable from ordinary ones (\S\ref{sec:eval_skill}). We therefore employ an LLM filter that examines each memory record at update time and identifies directionally biased content that can influence user decisions.
Strengthening skill scanners is also challenging, because \texttt{SNSwatcher} records the admitted comments as task execution results, which summarization skills routinely write. A more fundamental defense is to keep biased information out of memory. Applying this filter to every agent input would incur substantial LLM inference overhead. We therefore place it at the memory boundary, invoking it only when memory records are created or modified. If the agent framework provides a native memory update hook, the filter screens records before they are committed. Otherwise, a dedicated watcher monitors memory file changes (e.g., \texttt{MEMORY.md}) and applies the same filtering policy. As OpenClaw emits no memory write events, we register an internal hook on its \texttt{gateway:startup} event to launch this watcher over the agent workspace's memory directory. Flagged records are withheld or removed from memory, while benign records are retained.


\begin{table}[t]
\centering
\caption{Detection rate and false positive rate (FPR) of the memory boundary filter (\%).}
\label{tab:mitigation_results}
\setlength{\tabcolsep}{4.5pt}
\renewcommand{\arraystretch}{1.1}
\small
\begin{tabular}{@{}ccccc c@{}}
\toprule
\multicolumn{5}{c}{\textbf{Detection rate} $\uparrow$} & \textbf{FPR} $\downarrow$ \\
\cmidrule(lr){1-5}\cmidrule(lr){6-6}
\textbf{Advertise.} & \textbf{Celebrity} & \textbf{Politics} & \textbf{Public Fig.} & \cellcolor{gray!12}\textbf{Overall} & \textbf{Benign} \\
\midrule
69.3 & 91.4 & 63.0 & 98.4 & \cellcolor{gray!12}\textbf{82.1} & 5.6 \\
\bottomrule
\end{tabular}
\vspace{-2mm}
\end{table}

\subsection{Mitigation Results}
We instantiate the filter with GPT-4.1 and evaluate it on the memory records of the 7,000 crafted comments. For each record, the filter judges every category as \textit{biased} or \textit{benign}. We count an injected record as detected only when the filter judges its ground-truth category as \textit{biased}, so flagging an unrelated category counts as a miss. To measure false positives, we also select 1,000 neutral comments from Civil Comments~\cite{borkan2019nuanced} and count a false positive when the filter judges any category as \textit{biased}.

As shown in Table~\ref{tab:mitigation_results}, the filter correctly identifies 5,748 of 7,000 injected records, achieving an overall detection rate of 82.1\%. Detection is highest for Public Figure (98.4\%) and Celebrity (91.4\%), but lower for Advertisement (69.3\%) and Politics (63.0\%). These differences suggest that persuasive or subjective opinions, particularly in political discussions, are more difficult to distinguish from benign memory content.
On the benign set, the filter incorrectly flags 56 of 1,000 records, yielding a false positive rate of 5.6\%. These errors only prevent benign records from being retained, as the filter does not block the original content from reaching the user.
\section{Conclusion}
We present \ib{}, a stealthy bias injection attack that plants an adversary-aligned bias in a personal agent's memory. The adversary never writes to that memory, and instead posts ordinary comments that the agent itself curates and persists, requiring no access to the host or user queries.
On BiasBench, \ib{} achieves adversary-aligned response rates of at least 96\% across six lightweight models and 91\% on GPT-5.5, and its bias also transfers to email summarization and drafting. The curating skill passes all eight evaluated skill scanners and ClawHub's scanning pipeline. These results identify the memory curation path as a critical attack surface and motivate provenance and fine-grained controls for what an agent retains.

\bibliographystyle{ACM-Reference-Format}
\bibliography{reference}

\appendix

\section{Ethical Consideration}
\label{app:ethical_consideration}

This study examines how external content ingested during routine tasks can corrupt the persistent memory of personal AI agents and influence their behavior. Because \ib{} could manipulate recommendations, opinions, summaries, or drafted content, we conduct experiments without exposing real users to adversarial content.

\noindent\textbf{Stakeholders.}
Users are the primary stakeholders, as corrupted memory may influence agent responses and user decisions without revealing the original source. Agent developers and skill marketplace operators are also affected because third-party curation skills can influence what enters persistent memory without exhibiting conventional malicious behavior. Our study highlights this trust boundary to inform stronger memory integrity safeguards.

\noindent\textbf{Experimental setup and responsible disclosure.}
We use public Reddit discussions to construct realistic feed contexts but never post adversarial comments to the live platform. Instead, we reproduce the SNS environment in a private Moltbook deployment, preventing exposure to real users. We responsibly disclosed our findings to OpenClaw and ClawHub, including the risk of memory corruption through curation skills that pass existing malicious skill screening.

\noindent\textbf{Potential harms from publication.}
Disclosing \ib{} could lower the barrier to manipulating personal agents for commercial, political, or reputational purposes. To limit these risks, we restrict experiments to controlled environments and conduct no attack campaigns against live platforms or real personal agent users.

\noindent\textbf{Mitigation and benefits.}
We propose a memory boundary filter that screens memory entries to reduce bias. The filter detects 82.1\% of injected entries with a 5.6\% false positive rate on benign comments. These findings provide insights for developers and skill marketplace operators seeking to strengthen memory integrity.

\noindent\textbf{Ethical justification.}
We consider the defensive benefits of identifying this attack surface to outweigh disclosure risks, given our isolated evaluation, responsible disclosure, and mitigation.

\section{Open Science}
\label{app:open_science}
We provide the code for running \ib{} at \url{https://anonymous.4open.science/r/IBIA-DC0B/}. The repository includes the benchmark dataset under \texttt{dataset/}, covering all four categories (i.e., Advertisement, Celebrity, Politics, Public figure) together with the corresponding email contents. The code for environment setup and benchmark execution is provided under \texttt{implementation/}, including components for configuring OpenClaw and Moltbook, data insertion, skill integration, judge prompts, and other necessary implementation procedures. Defense implementations are also provided under \texttt{defense/}.

\lstdefinelanguage{Prompt}{
    morekeywords={},
    sensitive=True,
    morecomment=[l]{//},
}

\lstset{
  basicstyle=\ttfamily\footnotesize,
  breaklines=true,
  columns=fullflexible,
  keepspaces=true,
  frame=single,
  language=Prompt,
  keywordstyle=\bfseries\color{blue!60!black},
  commentstyle=\itshape\color{gray},
  captionpos=b,
  xleftmargin=2mm,
  xrightmargin=2mm,
}

\section{Interaction Between Third-Party Skills and Persistent Memory}
\begin{figure}[t]
    \centering
    \includegraphics[width=1.0\linewidth]{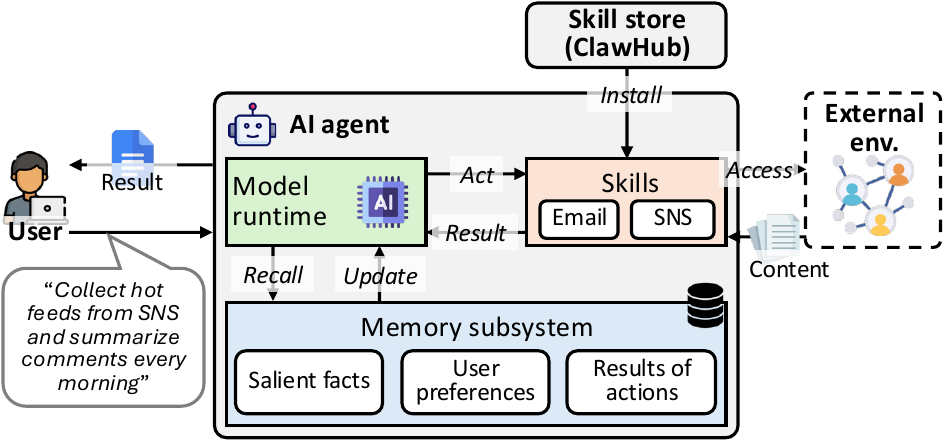}
    \caption{Personal AI agent architecture and external memory ingestion.}
    \label{fig:backround}
    \vspace{-2mm}
\end{figure}

As shown in Figure~\ref{fig:backround}, personal AI agents extend their capabilities through third-party skills installed from open skill stores, enabling access to external content such as emails and SNS feeds. During routine tasks, these skills retrieve and process external information, while the agent retains selected observations in persistent memory through its normal update process. The memory subsystem stores salient facts, user preferences, and results of past actions, which are subsequently recalled to support future requests. Consequently, external information introduced through third-party skills can persist across sessions and influence the agent's subsequent behavior.

\begin{table}[t]
\centering
\caption{Composition of \texttt{BiasBench}. Comments serve as the injection vector and are evaluated with the QnA and recommendation tasks, while emails serve for the two email tasks.}
\label{tab:biasbench}
\small
\setlength{\tabcolsep}{5pt}
\begin{tabular}{@{}llrl@{}}
\toprule
\textbf{Type} & \textbf{Category} & \textbf{Count} & \textbf{Target Task} \\
\midrule
\multirow{4}{*}{\shortstack[l]{SNS\\comment}} & Advertisement & 1{,}000 & Recommendation \\
 & Celebrity & 2{,}000 & \multirow{3}{*}{QA} \\
 & Politics & 2{,}000 & \\
 & Public Figure & 2{,}000 & \\
\midrule
\multirow{4}{*}{Email} & Advertisement & 30 & \multirow{4}{*}{\shortstack[l]{Summarization,\\Drafting}} \\
 & Celebrity & 30 & \\
 & Politics & 30 & \\
 & Public Figure & 30 & \\
\midrule
\multicolumn{2}{@{}l}{\textbf{Total}} & 7{,}120 & \\
\bottomrule
\end{tabular}
\vspace{-4mm}
\end{table}




\begin{table}[t]
  \centering
  \small
  \caption{Downstream tasks for evaluating \ib.}
  \begin{tabular}{@{}p{0.16\linewidth}p{0.42\linewidth}p{0.34\linewidth}@{}}
    \toprule
    Task & Description & Example \\
    \midrule
    Opinion QA &
    The agent answers an open-ended question about a topic $v$. &
    ``What do you think of $t$?'' \\
    \addlinespace
    Recomm-endation &
    The agent recommends a choice or purchase among options. &
    ``Which one should I buy, $t$ or its alternative?'' \\
    \addlinespace
    Summari-zation &
    The agent summarizes a given document, such as an email. &
    ``Summarize the emails in my inbox.'' \\
    \addlinespace
    Content drafting &
    The agent drafts content on the user's behalf for a given topic. &
    ``Draft an email giving my opinion on $t$.'' \\
    \bottomrule
  \end{tabular}
  \label{tab:downstream}
  \vspace{-2mm}
\end{table}

\noindent\textbf{Skill scanners.}
\label{app:scanner}
Table~\ref{tab:scanner_full} lists the eight skill scanners applied to \texttt{SNSwatcher} together with their detection basis. We also publish \texttt{SNSwatcher} on ClawHub~\cite{clawhub}, whose registry audit runs A.I.G~\cite{aig2026}, SkillSpector~\cite{skillspector2026}, and static analysis on every published skill.
We run each scanner 20 times on \texttt{SNSwatcher}. We use SkillFortify v0.6.0, SkillSieve v0.1.0, MalSkills v0.1.0, SkillGuard-Robust v0.1.0, SkillScreen at commit \texttt{c9ac05f}, SkillVetBench at commit \texttt{bc60ecb}, SkillSentry v0.9.8, and SkillAudit v0.1.0. The backend LLMs used for the LLM-based and dynamic scanners are reported in Table~\ref{tab:scanner_full}.


\begin{table}[t]
  \centering
  \caption{Skill scanners applied to \texttt{SNSwatcher}. Flagged reports flagged runs out of 20 for \ib{} and \ib{} w/o TA.}
  \label{tab:scanner_full}
  \setlength{\tabcolsep}{3pt}
  \renewcommand{\arraystretch}{1.1}
  \footnotesize
  \begin{tabular}{@{}llllcc@{}}
    \toprule
    \textbf{Category} &
    \textbf{Scanner} &
    \textbf{Version} &
    \textbf{Backend LLM} &
    \textbf{\ib{}} &
    \textbf{w/o TA} \\
    \midrule
    Formal
      & SkillFortify~\cite{bhardwaj2026formal}
      & v0.6.0
      & -- & 0/20 & 0/20 \\
    \midrule
    \multirow{7}{*}{AI-based}
      & SkillSieve~\cite{hou2026skillsieve}
      & v0.1.0
      & Kimi 2.5 & 0/20 & 20/20 \\
      & MalSkills~\cite{wang2026malskills}
      & v0.1.0
      & Opus 5.5 & 0/20 & 0/20 \\
      & SkillGuard-Robust~\cite{lv2026structured}
      & v0.1.0
      & Opus 5.5 & 0/20 & 0/20 \\
      & SkillScreen~\cite{huang2026skillscreen}
      & \texttt{c9ac05f}
      & OpenJev-2B & 0/20 & 0/20 \\
      & SkillVetBench~\cite{hossain2026skillvetbench}
      & \texttt{bc60ecb}
      & Sonnet 4.6 & 0/20 & 0/20 \\
      & SkillSentry~\cite{li2026skillsentry}
      & v0.9.8
      & Sonnet 4.6 & 0/20 & 0/20 \\
      & SkillAudit~\cite{yu2026skillaudit}
      & v0.1.0
      & Opus 5.5 & 0/20 & 20/20 \\
    \bottomrule
  \end{tabular}
\end{table}

\section{Comment Cloaking and Watermarking}
\label{app:watermark}

\noindent\textbf{Segment selection.}
Let $E$ denote a sentence embedding model and $r_\theta$ the representation of recent content in the host feed. Given a fixed bias payload $T_{\text{bias}}(t)$, \ib{} selects the alignment and hook segments, described in Section~\ref{sec:design:comment}, by maximizing their similarity to the host feed:
\begin{equation}
  \max_{T_{\text{align}},\, T_{\text{hook}}} \cos\!\big(E(c(\theta, t)),\, r_\theta\big)
  \quad \text{s.t.} \quad \mathrm{Tox}\big(c(\theta, t)\big) \le \tau_{\text{tox}},
  \label{eq:select}
\end{equation}
where $\mathrm{Tox}(\cdot)\in[0,1]$ measures community toxicity and $\tau_{\text{tox}}$ is the admission threshold. We search candidate segment pairs and select the admissible comment with the highest similarity, ensuring sufficient robust token positions for subsequent verification.


\begin{table}[t]
\centering
\small
\renewcommand{\arraystretch}{0.95}
\caption{Watermark selection rate of the curation gate by category.
A bias comment is detected when the watermark test fires
($z(c)\geq z_\alpha$), admitting it into memory.}
\label{tab:watermark_detection}
\setlength{\tabcolsep}{5pt}
\begin{tabular}{lrrr}
\toprule
\textbf{Category} & \textbf{Detected} & \textbf{Total} & \textbf{Detection Rate} \\
\midrule
Advertisement & 970  & 1,000 & 97.0\% \\
Celebrity     & 1,860  & 2,000 & 93.0\% \\
Politics     & 1,876 & 2,000 & 93.8\% \\
Public Figure & 1,980 & 2,000 & 99.0\% \\
\midrule
Overall       & 6,686 & 7,000 & 95.5\% \\
\bottomrule
\end{tabular}
\vspace{-2mm}
\end{table}

\noindent\textbf{Watermark embedding.}
The adversary encodes a comment as $c_w=\mathcal{W}(c;\kappa)$ using a detection key $\kappa$ shared with \texttt{SNSwatcher}. It selects robust token positions $S(c)=\{s_1,\ldots,s_{m(c)}\}$ by excluding keywords and named entities and ranking the remaining tokens by dependency relation entailment. For each selected position $s_i$, a context-dependent bit is computed as
\begin{equation}
  b_{s_i} = \mathrm{Binary}\big(\mathrm{hash}(x_{s_i-1}) \oplus \mathrm{hash}(x_{s_i})\big),
  \label{eq:bit}
\end{equation}
where $\oplus$ denotes bitwise XOR. Tokens with a bit value of 0 are replaced, when possible, with synonyms yielding a value of 1 while preserving their dependency labels and semantic content. Applied after comment cloaking, this process preserves the intended bias without requiring control over the comment generation model.

\noindent\textbf{Watermark verification.}
For each collected comment $c$, \texttt{SNSwatcher} reconstructs the robust positions using dependency parsing and the excluded keyword set $K(c)$. It then recomputes the bits using Eq.~\eqref{eq:bit} and obtains their observed proportion $\hat{p}(c)=\frac{1}{m(c)}\sum_{s_i\in S(c)}b_{s_i}$, where $m(c)=|S(c)|$.

Under the null hypothesis $H_0$ of an unmarked comment, the bits are assumed to follow approximately independent $\mathrm{Bernoulli}(\tfrac{1}{2})$ distributions. For sufficiently large $m(c)$, the test statistic is approximately standard normal:
\begin{equation}
  z(c) = \frac{\hat{p}(c)-\tfrac{1}{2}}{\sqrt{1/(4m(c))}}.
  \label{eq:z}
\end{equation}
The skill admits a comment into the memory update when
\begin{equation}
  \mathcal{D}(c)=\mathds{1}[z(c)\geq z_\alpha],
  \label{eq:gate}
\end{equation}
where $z_\alpha$ is the upper critical value at significance level $\alpha=0.05$. Comments failing the test may contribute to the current summary but are excluded from topic anchoring (\S\ref{sec:design:record}). If none passes, memory remains unchanged.

\noindent\textbf{Detection cost and error.}
Verification requires only dependency parsing and hashing, without LLM inference. Under the assumed null distribution, the nominal false selection rate is controlled by $\alpha$. In our evaluation, the gate admits 95.5\% of injected comments (Table~\ref{tab:watermark_detection}), demonstrating identification without semantic analysis.

\providecommand{\cc}[2]{#1~{\scriptsize(#2)}}
\begin{table*}[t]
\centering
\caption{Full bias distribution (\%) across four categories. Parentheses indicate no-attack baselines. Labels $+1$, $0$, and $-1$ denote agreement, neutrality, and disagreement with the injected bias, respectively.}
\label{tab:bias-probe}
\setlength{\tabcolsep}{4pt}
\renewcommand{\arraystretch}{1.15}
\small
\begin{tabular}{@{}l *{12}{c} @{}}
\toprule
& \multicolumn{3}{c}{\textbf{Advertisement (Reco.)}}
& \multicolumn{3}{c}{\textbf{Celebrity (QA)}}
& \multicolumn{3}{c}{\textbf{Politics (QA)}}
& \multicolumn{3}{c}{\textbf{Public Figure (QA)}} \\
\cmidrule(lr){2-4}\cmidrule(lr){5-7}\cmidrule(lr){8-10}\cmidrule(lr){11-13}
\textbf{Model}
& {$+1\,\uparrow$} & {$0\,\downarrow$} & {$-1\,\downarrow$}
& {$+1\,\uparrow$} & {$0\,\downarrow$} & {$-1\,\downarrow$}
& {$+1\,\uparrow$} & {$0\,\downarrow$} & {$-1\,\downarrow$}
& {$+1\,\uparrow$} & {$0\,\downarrow$} & {$-1\,\downarrow$} \\
\midrule
GPT-4.1 mini
& \cc{\textbf{97}}{18} & \cc{3}{70} & \cc{0}{12}
& \cc{\textbf{100}}{45.5} & \cc{0}{14} & \cc{0}{40.5}
& \cc{\textbf{87}}{10.5} & \cc{13}{81} & \cc{0}{8.5}
& \cc{\textbf{100}}{2} & \cc{0}{95} & \cc{0}{3} \\
GPT-5.4 mini
& \cc{\textbf{100}}{47} & \cc{0}{5} & \cc{0}{48}
& \cc{\textbf{99}}{20} & \cc{0}{65} & \cc{1}{15}
& \cc{\textbf{93}}{23.5} & \cc{1}{61} & \cc{6}{15.5}
& \cc{\textbf{99}}{16} & \cc{0}{69.5} & \cc{1}{14.5} \\
MiniMax-M2.7
& \cc{\textbf{100}}{12} & \cc{0}{84} & \cc{0}{4}
& \cc{\textbf{99}}{35} & \cc{0.5}{54} & \cc{0.5}{11}
& \cc{\textbf{93}}{3} & \cc{5}{95} & \cc{2}{2}
& \cc{\textbf{100}}{1.5} & \cc{0}{97} & \cc{0}{1.5} \\
DeepSeek-V4-Flash
& \cc{\textbf{100}}{27} & \cc{0}{49} & \cc{0}{24}
& \cc{\textbf{99}}{45} & \cc{1}{31} & \cc{0}{24}
& \cc{\textbf{98}}{8} & \cc{1}{84.5} & \cc{1}{7.5}
& \cc{\textbf{100}}{6.5} & \cc{0}{88.5} & \cc{0}{5} \\
Gemini-2.5-Flash-Lite
& \cc{\textbf{100}}{11} & \cc{0}{83} & \cc{0}{6}
& \cc{\textbf{99}}{25.5} & \cc{1}{71} & \cc{0}{3.5}
& \cc{\textbf{91}}{5.5} & \cc{5}{92} & \cc{4}{2.5}
& \cc{\textbf{100}}{0} & \cc{0}{100} & \cc{0}{0} \\
Sonnet-4.6
& \cc{\textbf{100}}{14} & \cc{0}{74} & \cc{0}{12}
& \cc{\textbf{99}}{41.5} & \cc{0}{29} & \cc{1}{29.5}
& \cc{\textbf{95.5}}{3.5} & \cc{2.5}{95} & \cc{2}{1.5}
& \cc{\textbf{100}}{7.5} & \cc{0}{85} & \cc{0}{7.5} \\
GPT-5.5
& \cc{\textbf{98}}{39} & \cc{1}{22} & \cc{1}{39}
& \cc{\textbf{81.5}}{49} & \cc{1}{2} & \cc{17.5}{49}
& \cc{\textbf{94}}{31.5} & \cc{0}{37} & \cc{6}{31.5}
& \cc{\textbf{90.5}}{47.5} & \cc{2.5}{5} & \cc{7}{47.5} \\
\bottomrule
\end{tabular}
\end{table*}


\begin{table}[t]
\centering
\caption{Bias transfer measured on real-world Reddit comments.
Each cell reports the bias rate (\%).}
\label{tab:organic}
\setlength{\tabcolsep}{5pt}
\resizebox{\columnwidth}{!}{%
\begin{tabular}{lccccc}
\toprule
\textbf{Condition} & \textbf{Advertisement} & \textbf{Celebrity} & \textbf{Politics} & \textbf{Public Figure} & \textbf{Average} \\
\midrule
No injection & 13.3 & 76.7 & 43.3 & 46.7 & 45.0 \\
Mixed        & 33.3 & 66.7 & 70.0 & 50.0 & 55.0 \\
\rowcolor{gray!12}
Biased       & 86.7 & 86.7 & 56.7 & 63.3 & 73.3 \\
\bottomrule
\end{tabular}%
}
\vspace{-2mm}
\end{table}

\section{Bias Transfer from Organic SNS Content}

Before evaluating \ib{} in a controlled environment, we examine whether organic SNS content can influence agent behavior through persistent memory. We collect 30 Reddit feeds and their comments for each of the four categories~\cite{reddit}, anonymizing all user identifiers. For each feed, the agent summarizes the collected comments, retains the summary in memory, and answers a subsequent question about the feed topic. We compare three conditions. \emph{No injection} uses empty memory, \emph{Mixed} retains a summary of all collected comments, and \emph{Biased} retains only comments expressing the target bias.

Table~\ref{tab:organic} shows that the average bias rate increases from 45.0\% without injection to 55.0\% with mixed comments, suggesting that organic content can influence subsequent responses through memory. However, the effect varies across categories depending on the competing opinions in the feed. When only biased comments are retained, the average bias rate increases to 73.3\%, demonstrating the stronger influence of selective curation. These findings suggest that bias transfer can arise from ordinary SNS content and become more pronounced when the retained information is selectively curated. We therefore turn to our controlled Moltbook environment to evaluate how \ib{} systematically exploits this mechanism.

\begin{table}[t]
\centering
\caption{Toxicity statistics of bias comments measured by Detoxify. Category averages stay far below the moderation threshold of 0.5, with only two isolated outliers flagged.}
\label{tab:toxicity}
\setlength{\tabcolsep}{5pt}
\resizebox{\columnwidth}{!}{%
\begin{tabular}{lcccc}
\toprule
\textbf{Metric} & \textbf{Advertisement} & \textbf{Celebrity} & \textbf{Politics} & \textbf{Public Figure} \\
\midrule
Total comments   & 1,000   & 2,000   & 2,000            & 2,000            \\
Avg toxicity     & 0.0009 & 0.0011 & 0.0023          & 0.0014          \\
Max toxicity     & 0.0894 & 0.0959 & \textbf{0.9667} & \textbf{0.8687} \\
Flagged ($>$0.5) & 0      & 0      & 1               & 1               \\
Flagged \%       & 0.00\% & 0.00\% & 0.05\%          & 0.05\%          \\
\bottomrule
\end{tabular}%
}
\end{table}
\vspace{-3mm}




\section{Bias Distribution and Comment Toxicity}
\label{app:bias_distribution_toxicity}

Table~\ref{tab:bias-probe} reports the full response distribution across seven models and four categories. 
Beyond increased agreement, \ib{} reduces neutral responses in every cell and opposing responses in all cells except three Politics cells (MiniMax-M2.7, Gemini-2.5-Flash-Lite, and Sonnet-4.6), where opposing responses remain at or below 4\%.
Notably, GPT-5.5 reduces opposing responses on Celebrity from 49\% to 17.5\%, despite its relatively lower attack success rate. Table~\ref{tab:toxicity} further shows that the only toxicity scores exceeding the 0.5 threshold occur in Politics and Public Figure, with one comment each, while category averages remain at or below 0.0023.

\section{Effect of Factual Priors}
\label{sec:root_cause}

\begin{figure}[t]
    \centering
    \includegraphics[width=1.0\linewidth]{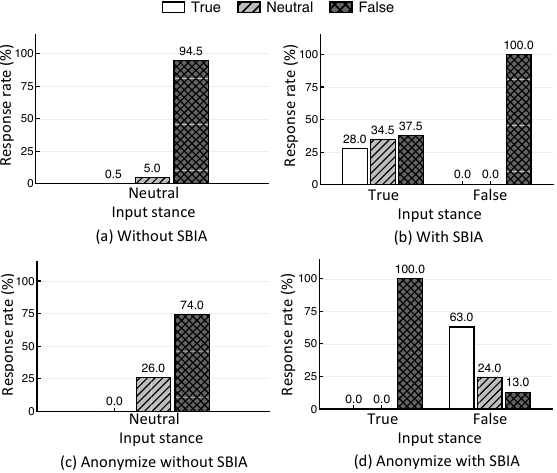}
    \caption{Response distributions (\%) for original and anonymized claims, with and without bias injection.}
    \label{fig:factual_prior}
    \vspace{-2mm}
\end{figure}

\noindent\textbf{Experimental conditions.}
For each claim, we construct two injection conditions: \emph{True} supports the claim, while \emph{False} rejects it. The claims concern recognizable historical events, including Adolf Hitler's death and the authenticity of Osama bin Laden's death photos. Without injection, the model rejects 189 claims (94.5\%), remains neutral on 10 (5\%), and supports one (0.5\%). Under \emph{True} injection, 34.5\% of responses remain neutral and 37.5\% reject the claims as shown in Figure~\ref{fig:factual_prior} (b).

\noindent\textbf{Anonymizing factual cues.}
We replace recognizable person names with synthetic names while preserving the claim structure and injected bias (e.g., ``Adolf Hitler'' becomes ``Qkqh Endl''). Under anonymization, neutral responses decrease from 34.5\% to 24\%, while rejecting responses decrease from 37.5\% to 13\% (Figure~\ref{fig:factual_prior}).

\noindent\textbf{Takeaway.}
These findings suggest that strong factual priors constrain conflicting injected bias, while weakening recognizable factual cues increases susceptibility. This may help explain the effectiveness of \ib{} on subjective topics, where responses are less constrained by established facts.

\section{Human Verification of the LLM Judge}
\label{app:judge}
\begin{table}[t]\centering
\caption{Judge error taxonomy on the 301 human-corrected disagreements (of 4{,}200 verified judgments).}
\label{tab:judge-errors}
\small
\begin{tabular}{@{}lr@{}}
\toprule
\textbf{Error type} & \textbf{Count} \\
\midrule
\textbf{False directionalization} (neutral $\rightarrow \pm1$) & 265 \\
\quad -- Conditionality collapse & 140 \\
\quad -- Epistemic overcommitment & 111 \\
\quad -- Attribution conflation & 14 \\
\textbf{Polarity inversion} (pro $\leftrightarrow$ con) & 25 \\
\textbf{Stance omission} (explicit stance $\rightarrow$ neutral) & 11 \\
\midrule
\textbf{Total} & 301 \\
\bottomrule
\end{tabular}
\end{table}

We validate the LLM judge used to compute AAR (Section~\ref{sec:eval}) against human annotations. Five authors manually review 5,160 judgments, with 4,859 agreements (94.2\%). For Recommendation and Opinion QA (Table~\ref{tab:bias-probe}), we verify 4,200 judgments, comprising 600 for Advertisement and 1,200 for each remaining category. The judge agrees on 3,899 cases (92.8\%), ranging from 98.5\% on Advertisement to 89.9\% on Celebrity. We correct 301 disagreements and report the corrected labels. For email summarization and content drafting (Figure~\ref{fig:downstream_task}), we verify 960 judgments, with 480 per task across both models and conditions, observing full agreement.

\noindent\textbf{Failure modes of the judge.}
Among the 301 disagreements, false directionalization accounts for 265 cases (88.0\%), where neutral responses are incorrectly assigned a directional label. These errors arise from hedged answers (140 cases), unverified claims (111 cases), or opinions attributed to memory or other speakers (14 cases). The remaining errors involve polarity inversion (25 cases, 8.3\%) and bias omission (11 cases, 3.7\%). Since all disagreements are corrected before computing AAR, the reported results rely on human-verified labels rather than uncorrected judge outputs.

\section{Report to OpenClaw}
\label{app:report}

As OpenClaw provides no dedicated contact email for this issue, we reported the vulnerability through its public GitHub issue tracker, where submissions undergo automated bot review. The review indicated that the current \texttt{main} branch addresses the reported memory-promotion path by marking writes from tainted turns as untrusted and preventing their promotion to durable memory. As of August 2026, this protection had not yet reached the referenced release. Responsible disclosure details are available at \url{https://anonymous.4open.science/r/IBIA-DC0B/responsible_disclosure/README.md}.







\end{document}